# Integrating Linguistics and AI: Morphological Analysis and Corpus development of Endangered Toto Language of West Bengal[1]

Ambalika Guha[1], Sajal Saha[1], Debanjan Ballav[1], Soumi Mitra[1], Hritwick Chakraborty[1]

Adamas University India[1]

**ABSTRACT**

Preserving linguistic diversity is necessary as every language offers a distinct perspective on the world. There have been numerous global initiatives to preserve endangered languages through documentation. This paper is a part of a project which aims to develop a trilingual (Toto-Bangla-English) language learning application to digitally archive and promote the endangered Toto language of West Bengal, India. This application, designed for both native Toto speakers and non-native learners, aims to revitalize the language by ensuring accessibility and usability through Unicode script integration and a structured language corpus. The research includes detailed linguistic documentation collected via fieldwork, followed by the creation of a morpheme-tagged, trilingual corpus used to train a Small Language Model (SLM) and a Transformer-based translation engine. The analysis covers inflectional morphology such as person-number-gender agreement, tense-aspect-mood distinctions, and case marking, alongside derivational strategies that reflect word-class changes. Script standardization and digital literacy tools were also developed to enhance script usage. The study offers a sustainable model for preserving endangered languages by incorporating traditional linguistic methodology with AI. This bridge between linguistic research with technological innovation highlights the value of interdisciplinary collaboration for community-based language revitalization.

## 1. Introduction

Toto, a Sino-Tibetan language, is domestically spoken by the Toto tribal community, belonging to the Particularly Vulnerable Tribal Groups (PVTGs) of India, residing in a village named Totopara in the Alipurduar district of West Bengal near the Indo-Bhutan border. In Totopara, communication takes place mostly in Nepali and Bengali to interact with other inhabitants of the neighbourhood. The official language of the Totopara is Bangla, which is the medium of instruction in schools. However, Nepali can be considered the lingua franca in Totopara (Piplai, 2018). With fewer than 1,700 Toto individuals, the language is facing a serious threat of extinction. Totos occupy themselves in wage labour along with farming on their land. Most of the Toto youths engage in contract-based work for which they travel to Nepal, Bhutan, Kolkata, Darjeeling, and other places. The linguistic and cultural vulnerability of Totos remains a constant challenge, henceforth. As language science is concerned, the preservation and revitalization of Toto is a fundamental requirement to deal with this threat of extinction. Toto possesses unique linguistic features that distinguish it from major Indo-Aryan languages like Nepali, Bangla, or Hindi, which are very prevalent in the region. Although, despite being a critically endangered ethnic community, and being a community of a small population, situated among such a linguistically and culturally diverse region, Toto manages to practice their ethnicity. However, it is required to preserve and conserve the language periodically to understand the changes taking place diachronically. In this case, the language of Totos remains one of the lesser-studied and under-documented in the Tibeto-Burman branch of the Sino-Tibetan family of languages.

---

[1] This research is conducted under the ICSSR-sponsored major project. The author(s) gratefully acknowledge the financial and institutional support provided by the Indian Council of Social Science Research (ICSSR), Government of India.

This paper is part of a broader ongoing project, funded by the ICSSR (Indian Council of Social Science Research), that aims to preserve and revitalize the Toto language through an interdisciplinary approach. This approach integrates traditional linguistic methods with AI-based tools. The intent of this project is to develop a trilingual language learning application (Toto-Bengali-English). This app will give access to the young Toto speakers (school and college students), who are proficient in Bangla (one of the official languages of the state), to learn English (an economically viable language) via their native language, Toto, and the official language Bangla. In this way, the Toto speakers can remain connected to their mother tongue, and also, this model can revolutionize multilingual education, especially in indigenous and minority communities. Further, this app can be used by non-native Toto speakers, who know English and Bangla and visit or will visit Totopara for research work, as tourists, and educators, to learn the Toto language through Bangla or English. In this way, the number of Toto language users might increase. In this app, the mediator language will be Bangla, which acts as the common language between the native and non-native Toto speakers. This app can be used as a digital primer for the young Toto speakers, with the help of which they can read the letters, words, phrases, sentences, and short conversational texts represented in the Toto language script. Moreover, this project will be the first attempt to generate various types of sentences and short conversational texts written in the Toto language script, which will be preserved in this app. The app will encourage both native and non-native Toto speakers to use the Toto language. Such efforts can be in line with policies aimed at promoting and revitalizing indigenous languages. This research is significant for several reasons. First, it directly contributes to the preservation of an endangered language by making it accessible to a broader audience, including young native speakers and non-native learners. Second, it promotes multilingual education, allowing Toto speakers to gain proficiency in English—a language of economic and social mobility—without abandoning their indigenous tongue. Third, it advances linguistic research by providing a structured, AI-enhanced dataset of the Toto language, contributing to the broader field of endangered language documentation. The project also aims to preserve the Toto script, an alphabetic script, which was developed by community elder and author Dhaniram Toto in the year 2015.

The objective of this paper is to provide a detailed morphological analysis of the Toto language that includes providing a structured account of inflectional and derivational morphemes. The study includes extensive field work, community collaboration, and corpus-building to offer a comprehensive understanding of the language's internal grammar. Unlike prior documentation efforts that remain static, descriptive, this research contributes to creating an annotated corpus that feeds directly into a Small Language Model (SLM), which is tailored for low-resource NLP environments. Altogether, this research aims to offer a model for endangered languages preservation that is scalable, accessible, and technologically forward-looking.

*1.1. Literature review*
*1.1.1. Language endangerment*
Language endangerment is a critical concern for many language communities, as language is often seen as a foundation of cultural identity. According to [12], a language is endangered when speakers stop using their native language, when intergenerational transmission breaks down, or when the use of the language becomes significantly reduced. The issue of language endangerment first gained attention within the international linguistic community during the 1970s. In 1992, Chinese linguists were formally introduced to the concept of endangered languages during an international congress [13]. During the 1990s, UNESCO began initiating efforts to document endangered languages globally. In 2000, two influential works on language endangerment were published: Language Death by David Crystal and Vanishing Voices: The Extinction of the World's Languages by [22] highlights that [28] Endangered Languages: An Introduction (2015) is considered the first textbook-style publication in the field. Several countries—including the United Kingdom, the United States, China, Korea, and India—as well as organizations such as UNESCO, UGC, CIIL, and SOAS University of London, have undertaken initiatives to document endangered languages. Academic institutions have contributed to awareness through seminars, films, workshops, and public outreach. In 1992, the Linguistic Society of America (MIT) published a series of works on endangered languages authored by [15]. These researchers examined the endangered languages of Arnhem Land and Cape York Peninsula in Australia, as well as the bilingual Sumú and Miskitu languages of Central America.

In India, several efforts have been made to preserve endangered tribal languages and cultures. The Great Andamanese languages, including Jarwa and Onge, were documented by [1- 3] as part of the Endangered Languages Documentation Project (Vanishing Voices of the Great Andamanese) coordinated by SOAS, University of London. Dr. Bikram Jora, regional coordinator for South Asia with the Living Tongues Institute, has worked extensively on documenting endangered languages such as Bhumij, Ho, Santali, Kharia, Kera' Mundari, and Tamari Mundari. He has also collaborated with Dr. Gregory Anderson to document endangered languages of Arunachal Pradesh, including Koro Aka, Hruso Aka, Bangru, Puroik, and Sartang.

Faculty members from Jadavpur University, Delhi University, and Calcutta University are also actively engaged in documenting the moribund languages of tribal communities in North Bengal, West Bengal, and the North-East region. The Central Institute of Indian Languages (CIIL), Mysore, has implemented projects under the Government of India's Scheme for Protection and Preservation of Endangered Languages. These projects focus on developing dictionaries, primers, grammatical sketches, and pictorial glossaries. The Ministry of Tribal Affairs has supported research on endangered languages through initiatives involving the creation of bilingual dictionaries, trilingual proficiency modules, rhymes, storybooks, and other educational materials. Under the scheme 'Tribal Research, Information, Education, Communication and Events (TRI-ECE)', the government has funded the development of AI-based translation tools to convert English and Hindi text or speech into selected tribal languages and vice versa. The Bhasha Research and Publication Centre in Vadodara has worked on a TRI-ECE-funded project focusing on the study and documentation of Adivasi languages, cultures, and life skills. Currently, BITS Pilani, in collaboration with a consortium of IITs. A major collaboration has been initiated by various organizations, including government agencies, startups, and academia, to contribute to the project Bhasini, a project that solely focuses on AI-based translation tools under the TRI-ECE scheme.

*1.1.2. Previous works on Toto*
In the past few years, some of the Toto community members, film directors, and writers have been intensely trying to preserve the language and culture of this critically endangered community. One of the significant steps towards the preservation of Toto language began in 2015 with the maiden development of its script by Dhaniram Toto. The film director, Rajaditya Banerjee, made a documentary, Lost for Words (2017), to sketch the struggle of this tribe to preserve their language and culture.

Linguistically and culturally, there are significant studies that have captured the descriptive nature of Toto grammar, including works of [10] that have explored Phonology, syllable structure, and Morpho-syntax of Toto. [9] in their short book 'A Descriptive Study of Toto Language' has detailed the phonetic inventory along with a description of Phonology, Morphological formation, and Syntax. [6] in his PhD thesis revisits the descriptive grammar of Toto. [24], [25] on the other hand, has focused on the ethnic life of Totos, presenting factors like settlement, social & domestic life, social organization, and their language. Perumal Samy (2016) documented Toto's phonology and morphology, highlighting case marking and verb typology. In 2019, Professor Atanu Saha from Jadavpur University worked on a UGC funded project to present a linguistic description of this language. In 2023, a printed dictionary of the Toto language was developed by a Toto community member, Bhakta Toto and a professor from Calcutta University, Mrinmoy Pramanik. We further learn better about the Toto community through the anthropo-linguistic research work by Dripta Piplai (2018) where she has brought the volatile nature of Toto morpho-syntax. This study is important as it periodically traces linguistic changes in Toto compared to their settlement (Piplai 2018). She highlights how Toto morphology has undergone severe changes in the last 50-60 years. Changing socio-cultural behavior of Toto tribes by Shahid Jamal, Aakash Upadhyay, and Rachna Dua (2021), and Toto tribe's livelihood pattern by Manas Mohan Adhikary et al. (2024) are also significant as the studies capture the current socio-cultural-economic condition of the Toto community.

Despite all previous significant contributions, in the era of modern technology and its incorporation with liberal studies and social sciences, it is important to consider taking a digital archiving approach for language documentation as well, especially for underrepresented languages.

*1.2. Research questions*
Based on the goals and objectives of this study, the following are some fundamental research questions this paper has focused on:
- What is the current state of inflectional and derivational morphological features of the Toto language? This question deals with Person-Number-Gender (PNG), Tense-Aspect-Mood (TAM), the case system, and their internal intricacies, and other peripheral morphemic identifications and overlapping.
- How can a morpheme-tagged, trilingual corpus (Toto-Bangla-English) be developed to train AI models like Small Language Models (SLM) and Transformer-based engines for translation and language learning?
- How effective is the Unicode-based standardized Toto script in enabling digital literacy and usability within the app interface?
- Can the combination of traditional linguistic documentation and modern AI tools provide a sustainable and replicable model for endangered language preservation?

**2. Method**
To effectively document and analyze the Toto language while developing an AI-powered trilingual language learning application, this project employs a multidimensional methodological approach, including linguistic fieldwork, computational data processing, and AI-driven language modeling to ensure both accuracy and applicability. For this particular descriptive analysis of grammatical morphemes of Toto as a part of this project, the methodology involves systematic data collection from native Toto speakers, morphological analysis of linguistic structures, and the creation of a parallel Toto-Bangla-English corpus to train an AI-based learning model. The following key methodological steps outline the research process.

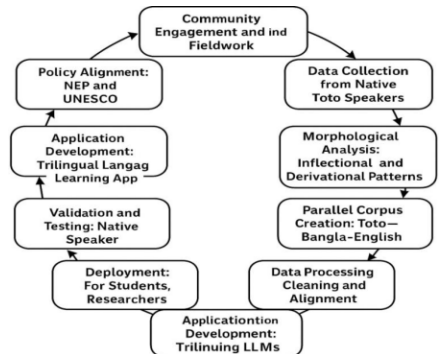

**Figure 1:** Research Methodology Applied in this Research

*2.1. Sample / Participants*
Based on the availability of the informants, the selection of participants for this study was done based on age groups. The age variable consists of regional male speakers under two age cohorts: those in their 20s–30s and those aged 60 and above, including students, private and government school teachers, and some retired males.

*2.2. Instrument(s)*
The documentation process typically includes a textual and audio-visual approach. According to the agenda of the project, the data in audio form becomes crucial, as the language learning application will consist of real speech. The textual content will be embedded with the recorded audio. To capture audio, a Zoom digital recorder was used. Secondly, a GoPro hand-cam was used to record some sessions.

With an aim to build a large corpus, the data was elicited with the help of a certain number of structured and unstructured questionnaires. The structured questionnaire mainly includes an exhaustive word list that was made based on different parts of speech, under which the words are organized in terms of their semantic category. For example, nouns and pronouns consisted of around 1500 words and 36 categories. This is the same procedure through which the lists of verbs, adjectives, and adverbs were made. The questionnaire for the lexicon list relied upon the Toto dictionary made by Bharat Toto and Mrinmay Pramanik (see literature review). The questionnaire made for inflection contains simple questions to capture noun morphology by documenting person, number, and gender agreements. For verb morphology, tense, aspect, and mood were captured. Also, a specific questionnaire was made to explore the case system in toto languages such including nominative, accusative, locative, genitive, dative, instrumental, and ablative.

Linguistic phenomena like nominalization, verbalization, and adjectivization were studied to understand the derivational morphology of Toto. In this part, the questionnaire contains certain parts of speeches from the lexicon list and how it changes to another category due to derivation, e.g., write > writer.

The unstructured questionnaire was made in the field, which includes certain context-based scenarios based on the observations. For example, how classifiers work in Toto, what kind of metaphorical extensions Toto shares, and how much Toto is influenced by Nepali or Bengali. Domains like time and space, kinship, motion, and directions can also be well understood from the conversation within a real-time discourse. In addition to that, the unstructured questionnaire was an extension of the structured questionnaire. In simple words, asking the same type of questions to different informants to understand the pattern is important, which may highlight further sociolinguistic variation among the Toto, which is not the focus of this study.

**3. Data collection procedures and analysis**
The field consisted of a month, during which around twenty days were devoted to data collection. All the data collection is in session form. The settings were chosen based on the informants' convenience. Most of the time, the setting would be indoor, for example, a homestay or private tuition. In cases of public places, mostly the data collection took place at a café run by a Nepali family. The sessions consisted of the engagement of three linguists and the informant.

The core elicitation procedures underwent several steps. While eliciting words, the informants were asked to repeat the term three times to understand and capture the pronunciation. It is important to capture available variations in order to understand the dominating linguistic traits of Toto. Every utterance by the informant was recorded by a Zoom audio recorder, and the sessions were also filmed by a GoPro hand-cam. The informants were also asked to maintain a pause of 8 seconds between each answer, so the trimming process takes less time while developing the application.

Engaging in discussions with the informant was an effort to document some oral narratives, despite conducting structured and organized sessions. This effort helped understand and record their interpretations on different topics. This built a 'natural environment' where the informant starts explaining by himself about any specific word, its different usages, or how a sentence can be constructed differently. These responses provide how Toto perceives the world. For example, the concepts of blood and milk can be expressed by a common term with a minor phonological difference. Or, the semantic notion of 'positive' and 'negative' under which opposites such as 'good' and 'bad', 'perfect' and 'imperfect' can be listed. All notions related to the semanticity of 'positive' and 'negative' can be represented by /eŋtaʋa/ and /meŋtaʋa/ respectively.

*3.1. Linguistic Analysis of Toto*
The paper presents the linguistic analysis of the Toto language in terms of inflectional and derivational morphemes. By categorizing lexical items based on their grammatical functions and semantic properties, this research contributes to a deeper understanding of Toto's linguistic framework. Inflectional morphemes are grammatical markers that indicate aspects such as tense, number, case, or degree. They do not change the word's core meaning or its word class (e.g., noun, verb, adjective). Derivational morphemes, on the other hand, create new words by adding prefixes or suffixes. They often change the meaning of the base word and sometimes its word class. This paper shows a detailed study of these two categories of morphemes in the critically endangered Toto language.

*3.1.1. Inflectional Morphemes*
This section of the paper introduces the inflectional morphemes of Toto language, broadly divided under the grammatical categories of Number, Tense and Aspect, Mood, Case, and Postpositions.

*3.1.1.1. Plural Morpheme in Toto*
Toto speakers use -bɪ morpheme to indicate plurality, cf. ((1)-(6)).

(1)  *kuŋ*          *ceŋ*  
     1SG.GEN       child.SG  
     'My child'

(2)  *kuŋ*          *ceŋ*       -*bɪ*  
     1SG.GEN       child        PL  
     'My children'

(3)  *kuŋ*          *pika*  
     1SG.GEN       cow.SG  
     'My cow'

(4)  *kuŋ*          *pika*      -*bɪ*  
     1SG.GEN       cow          PL  
     'My cows'

(5)  *kuŋ*          *ʈebil*  
     1SG.GEN       table.SG  
     'My table'

(6)  *kuŋ*          *ʈebil*     -*bɪ*  
     1SG.GEN       table        PL  
     'My tables'

The plural morpheme *-bi* is also used to indicate plurality of subject nouns, cf. 7. In (7) and (8), *-bi* is used to signal plurality in both subject nouns *ki-bi-kɔ* 'our' and *nətɪ-bɪ-kɔ* 'your' and the complement of subject nouns *ceŋ-bi* 'children'. The suffix *-kɔ* attached to the subject nouns in (7) and (8) is a genitive case morpheme, which will be discussed later in the case section of this paper.

(7)  *ki*       -*bɪ*      -*kɔ*       *Ceŋ*       -*bɪ*  
     1          PL          GEN         Child        PL  
     'Our children'

(8)  *nətɪ*     *bɪ*        *kɔ*        *Ceŋ*       -*bɪ*  
     2          PL          GEN         Child        PL  
     'Your children'

### 3.1.1.2. Tense and Aspect Morphemes in Toto

In Toto language, present, past, and future tenses appear to be marked with three distinctive morphemes *-mi* (present), *-na* (past), and *-ro* (future), as can be seen in (9), (10), and (11), respectively.

(9)     *ka*        *dinei*      *din*      *iskul*    *-ta*      *teipuːm*   *-ʃa*    *ha*    *-mi*
        1SG         every        day        school     LOC        walk        INS      go      PRS
        'I walk to school every day.'

(10)    *ka*        *ainɟi*      *iskul*    *-ta*      *teipuːm*  *-ʃa*       *ha*     *-na*
        1SG         yesterday    school     LOC        walk       INS         go       PST
        'I walked to school yesterday.'

(11)    *ka*        *ɟukuŋ*      *iskul*    *-ta*      *teipuːm*  *-ʃa*       *ha*     *-ro*
        I           tomorrow     school     LOC        walk       INS         go       FUT
        'I will walk to school tomorrow.'

However, it is further observed that the morphemes *-na* and *-mi* are sometimes used interchangeably for present and past tense. We have noticed in (9) and (10) that *-mi* and *-na* are used as present and past tense markers, respectively. Now, we will see in (12) and (13) that *-na* and *-mi* are used as present and past tense markers, respectively

(12)    *Ram*       *ʃəbri*      *Coi*      *-na*
        Ram         vegetables   Buy        PRS
        'Ram buys vegetables.'

(13)    *akɔ*       *ka*         *-hiŋ*     *naʃa*     *kicpa*    *-mi*
        3SG         Me           ACC        notes      lent       PST
        'She lent notes/money to me.'

As evident from the data in (9), (10), (12), and (13), it can be said that in Toto language, present tense and past tense are indicated by either *-na* or *-mi*. But the future tense morpheme is distinctively realized as *-ro*, cf. (11), (14), (15), and (16).

(11)    *ka*        *ɟukuŋ*      *iskul*    *-ta*      *teipuːm*  *-ʃa*       *ha*     *-ro*
        I           tomorrow     school     LOC        walk       INS         go       FUT
        'I will walk to school tomorrow.'

(14)    *Ka*        *iskuːl*     *-ta*      *teipuːm*  *-ʃa*      *ha*        *-ro*
        1SG         School       LOC        walk       INS        go          FUT
        'I will walk to school.'

(15)    *ka*        *ʃaːt*       *baɟi*     *-ta*      *canaŋ*    *ca*        *-ro*
        1SG         7            clock      LOC        breakfast  eat         FUT
        'I will eat breakfast at 7 o'clock.'

(16) ɟukuŋ   ape    -bi   park   -ta   kyalai   -ro
     tomorrow  Child  PL    park   LOC   play     FUT
     'Tomorrow, children will play in the park.'

The progressive aspect in Toto language is signaled by three different variations of *-daŋ* morpheme: *-daŋ, -diŋ, -duŋ*, as can be seen in (17) – (21). The case morphemes represented in the data given below will be discussed in detail in a later section.

(17) ka    nɛha   isku:l   -ta   teipu:m   -ʃa   ha   -daŋ   -na
     1SG   now    school   LOC   walk      INS   go   PROG   PRS
     'I am walking to school now.'

(18) ka    isku:l   -ta   teipu:m   -ʃa   ha   -daŋ   -mi
     1SG   school   LOC   Walk      INS   go   PROG   PST
     'I was walking to school.'

(19) ka    nɛha   hapuŋ    -ko    ama    ca:   -diŋ   -na
     1SG   now    morning  Gen    food   eat   PROG   PRS
     'I am now eating breakfast.'

(20) ka    hapuŋ    -ko    ama    ca:   diŋ    -mi
     1SG   Morning  GEN    food   Eat   PROG   PST
     'I was eating breakfast.'

(21) aku   bara    -ʃo    toiŋ   -duŋ   -na
     3SG   fence   ABL    jump   PROG   PRS
     'She/he is jumping over the fence.'

The perfect aspect in Toto is marked by *-pate* and *-pu* morphemes. Present and past perfect are indicated by *-pate* and the future perfect is signaled by *-pu* morpheme, as can be noticed in the data given below from (22)-(27).

(22) ka    i     iga    nagai    -ʃo    parai   -pate   -na
     1SG   DEM   book   before   ABL    read    PFV     PRS
     'I have read this book before.'

(23) mubi    waŋ-    waroma   nagai    ka     i     iga    parai   pate   -mi
     Movie   come    Part     before   1SG    DEM   book   read    PFV    PST
     'I had read this book before the movie was released.'

(24) klas    gin-     waroma   nagai    ʃo     ka     higa   parai   -pu    -ro
     class   start-   Part     before   ABL    1SG    book   read    PFV    FUT

'I will have read the book before the class starts.'

(25) | *ka* | *oise* | *bar* | *isku:l* | *-ta* | *teipu:m* | *ʃa* | *ha* | *-pate* | *-na* |
|---|---|---|---|---|---|---|---|---|---|
| 1SG | many | Times | School | LOC | walk | INS | go | PFV | PRS |

'I have walked to school many times.'

(26) | *boʃ* | *hadaŋwa* | *roma* | *nagai* | *ka* | *isku:l* | *-ta* | *teipu:m* | *-ʃa* | *ha* | *pate* | *-mi* |
|---|---|---|---|---|---|---|---|---|---|---|---|
| bus | service | start | before | 1SG | school | LOC | walk | INS | go | PFV | PST |

'I had walked to school before the bus service started.'

(27) | *nati* | *waŋ-* | *waroma* | *nagai* | *ka* | *isku:l* | *-ta* | *teipu:m* | *-ʃa* | *ha* | *pu* | *-ro* |
|---|---|---|---|---|---|---|---|---|---|---|---|
| 2SG | come- | part | before | 1SG | school | LOC | walk | INS | go | PFV | FUT |

'I will have walked to school before you arrive.'

*3.1.1.3. Mood in Toto*

In Toto, indicative mood is presented in simple present tense form (28) – (35).

(28) | *ako* | *hapta* | *-ta* | *pʰutbal* | *kelai* | *-mi* |
|---|---|---|---|---|---|
| 3SG | weekend | LOC | football | play | PRS |

'He/she plays football on the weekend.'

(29) | *ako* | *toi* | *-ʃo* | *toiŋ* | *-na* |
|---|---|---|---|---|
| 3SG | high | ABL | jump | PRS |

'She/he jumps from a height.'

(30) | *ako* | *dinei* | *porai* | *-na* |
|---|---|---|---|
| 3SG | everyday | read | PRS |

She/he reads every day

(31) | *ako* | *dinei* | *la* | *-mi* |
|---|---|---|---|
| 3SG | everyday | write | PRS |

'She/he writes every day.'

(32) | *ako* | *wŋtapa* | *ama* | *lei* | *-na* |
|---|---|---|---|---|
| 3SG | well | food | cook | PRS |

'She/he cooks well.'

(33) | *ako* | *ʃenepa* | *tui* | *-na* |
|---|---|---|---|
| 3SG | Fast | run | PRS |

'She/he runs fast.'

(34) | ako | eŋtapa | nui | -na
| 3SG | well | swim | PRS

'She/he swims well.'

(35) | ako | oiʃe | noaŋ | ɟaŋ | -na
| 3SG | lot | talk | do | PRS

'She/he talks a lot.'

Notice that in the above examples of indicative mood (28)-(35), the present tense is sometimes marked by -*na* morpheme and sometimes by -*mi* morpheme.

**Imperative mood** in Toto is reflected in the verb without any tense morpheme -*mi* and -*na*, as can be seen in (36)-(40).

(36) | bair | -ɛ | kelai
| outside | LOC | play

'Play outside.'

(37) | liʃuŋ- | -ko | ama | lei
| night | GEN | food | cook

'Cook dinner,'

(38) | ʃenepa | tui
| fast | run

'Rub fast.'

(39) | eŋtapa | nui
| fast | swim

'Swim fast.'

(40) | aku | -hiŋ | -pa | noaŋ | ɟe
| 3SG | ACC | to | talk | Do

'Talk to him.'

**Subjunctive mood** in Toto is indicated by the presence of the habitual form of the auxiliary -*ko* ((41) – (45)).

(41) | ka | pʰutbal | kelai | -na | ico | tim | -ta | kelai | -ko
| 1SG | football | play | PST | one | team | LOC | play | HAB

'If I had played football, I would have joined a team.'

(42)   ka        mi       toiŋ      -na       toi       -ta       Deka      -ko
       1SG       If       jump      PST       top       LOC       Reach     HAB
       'If I had jumped, I would have reached the top.'

(43)   ka        iga       porai     -na       hoaŋ      ce                  -ko
       1SG       book      read      PST       topic     understand          HAB
       'If I had read a book, I would have understood the topic.'

(44)   ka        mi        Ama       lei       -na       Ka        pʰasta    banai     -ko
       1SG       If        food      cook      PST       1SG       Pasta     make      HAB
       'If I were to cook, I would make pasta.'

(45)   ka        mi        Noaŋ      ɟena      eŋtapa    juŋ       -ko
       1SG       If        talk      were      careful   Be        HAB
       'If I were to talk, I would be careful.'

*3.1.1.4. Case Morphemes in Toto*
In Toto language, the following nominal cases could be identified: nominative, accusative, dative, genitive, locative, instrumental, and ablative.

**Nominative case** is presented by a null morpheme, as can be seen in (46)-(50).

(46)   ʃema      -Ø        ɣəbri      Coi       -na
       Shyam-    NOM       vegetables Buy       PRES
       'Shyam buys vegetables.'

(47)   ape       -bi       -Ø        kelai     -na
       Child     PL        NOM       Play      PRS
       'Children play.'

(48)   aku       -Ø        eŋtapa    Le        -na
       she/he-   NOM       beautifully sing    PRS
       'She sings beautifully.'

(49)   ɟiha      -bi       -Ø        hapuŋ     cecoi     -mi
       bird      Pl        NOM       morning   chirp     PRS
       'Birds chirp in the morning.'

(50)   ape       -bi       -Ø        poɾaim    -ʃa       seti-              -mi
       Student   Pl        NOM       study     PRS       exam               for
       'Students study for exam.'

**Accusative case** in Toto is marked by three morphemes: *-hẽ*, *-hiŋ*, *-hi*, as can be observed in (51) - (52) respectively.

(51) *kũa*   -bi   -Ø   *a*   -bi   -hiŋ   *ʃai*   -na
tiger   PL   NOM   3SG   PL   ACC   hunt   PRS
'Tigers hunt them.'

(52) *nars*   -bi   -Ø   *haim*   -bi   -hi   *eŋtapa   ɟɔ*   -mi
nurse   PL   NOM   patient   PL   ACC   care   do   -PRS
'Nurses care for patients.'

**Genitive case** in Toto is marked by *–ko* and *–kɔ* morpheme, cf. ((53) – (55)).

(53) *akɔ*   -Ø   *megevaɟiniʃi*   -ko   *gepa*   -na
3SG   NOM   Puzzle   GEN   solve   PST
She/he solved puzzle

(54) *aku*   -Ø   *Ico   bəlua*   -ko   *ʃa   banai*   -na
3SG   NOM   one   sand   GEN   castle   build   PST
She/he built a sandcastle

(55) *gita   sita*   -kɔ   *baɟeɾo*
Gita   Sita   GEN   friend
Gita is Sita's friend.

**Locative case** in Toto is marked by two types of morphemes: *-ta* and *-ʃo,* cf. (56) – (59).

(56) *cabi   tɔ*   -kɔ   *li*   -ta   *ni*   -na
key   mat   GEN   below   LOC   be   PRS
'Keys are below the mat.'

(57) *akɔ*   -Ø   *ʃinge*   -kɔ   *li*   -ta   *dʒa*   -mi
3SG   NOM   tree   GEN   below   LOC   stand-   PST
'She/he stood under the tree.'

(58) *kija*   -bi   *me   kɔ   aipi*   -ʃo   *ɟiŋ*   -diŋ   -na
dog   PL   Fireplace   GEN   beside   LOC   sleep   PROG   PRS
'Dogs are sleeping beside a fireplace.'

(59) *a*   -bi   *pɔkɔɾi*   -ko   *aipi*   -ʃo   *tei*   -mi
3   PL   lake   GEN   beside   LOC   walk   PST
'They walked beside a lake.'

**Dative case** in Toto is marked by two types of morphemes: *-hiŋ* and *-ta,* cf. (60) – (62). One interesting thing to note here is that in the case of ditransitive constructions, the direct object, which is supposed to be marked as accusative case, loses accusative case morphemes, and the indirect object, which has dative case, is marked with case morphemes.

(60)   *ɾam*   -Ø   *ka*   *-hiŋ*   *igɔ*   - Ø   *pica*   *-na*
       Ram    NOM   1SG   DAT      book    ACC   give     PST
       'Ram gave a book to him.'

(61)   *akɔ*   -Ø   *akɔ*   *-ta*   *cabi*   -Ø   *pica*   *-na*
       3SG    NOM   3SG    DAT     keys     ACC   give    PST
       'She/he gave keys to him/her.'

(62)   *akɔ*   *-bi*   -Ø   *abi*   *-ta*   *maibe*   -Ø   *kaipu*
       3SG    PL      NOM   3PL    DAT     flowers   ACC   Send
       'They sent flowers to them.'

Both **ablative** and **instrumental case** in Toto is marked with *-ʃɔ* morpheme, cf. ((63) – (66)).

(63)   *kuŋ*           *kɔlɔm*     *-ʃo*     *la*        *-ko*
       1SG.GEN         pen         INST      write       IMP
       'Write with my pen.'

(64)   *keŋci*         *-ʃo*       *iga*     *ca:*
       scissor         INST        book      Cut
       'Cut the book with scissors.'

(65)   *aku*           *ʃiŋge*     *-ʃo*     *nui*       *-na*
       3SG             tree        ABL       Fall        PST
       'He/she fell from the tree.'

(66)   *dal*           *-ʃo*       *lapa*    *ɟo*        *-mi*
       Branch          ABL         Leaves    fall        PST
       'Leaves fell from the branch.'

Please note, cf (63) *-ko* morpheme occurs as a suffix; however, it is playing the role of imperative mood. Such instances can be noticed in Toto; however, the imperative mood marker is not obligatory. In our upcoming research on this project, we will encounter such linguistic nuances that ultimately signify the volatile nature of Toto grammar.

*3.1.1.5. Morphemes -ha and -he in Toto*
Toto speakers use the morpheme -ha, attached to the nominal, to indicate definiteness, cf. ((67) – (69)).

(67)    *miŋki*    *-ha*    *oi*    *ɟuia*    *-hẽ*    *gʰi*    *-na*

       cat    DEF    that    mouse    ACC    chase    PST

       'The cat chased that mouse.'

(68)    *kũa*    *-ha*    *abi*    *-hiŋ*    *ʃai*    *-na*

       Dog    DEF    3SG    ACC    hunt    PRS

       'The tiger hunts it.'

(69)    *tʃabi*    *-ha*    *tɔ*    *-kɔ*    *li*    *-ta*    *ni*    *-na*

       keys-    DEF    mat    GEN    below    LOC    be    PRS

       'The keys are below a mat.'

The morpheme *-he* can be considered as an emphatic marker for the time being. In the examples given below (70)-(72), the occurrence of *-he* puts phonetic stress on the verb. More detailed study about *-he* is left for future study.

(70)    *a*    *-bi*    *abi*    *kɔ*    *ʃai*    *lɔŋ*    *lagai*    *-na*    *-he*

       3    PL    3PL    GEN    house    color    put    PST    EMPH

       'They painted their house.'

(71)    *ako*    *ico*    *cubə*    *gəɾi*    *ʧoi*    *-na*    *-he*

       3SG    one    New    car    buy    PST    EMPH

       'He bought a new car.'

(72)    *akɔ*    *iga*    *-ha*    *pəɾai*    *-na*    *-he*

       3SG    book    DEF    read    PRS    EMPH

       'She reads the book.'

### 3.1.2. Derivational Morphemes in Toto

By definition, derivational morphemes are bound morphemes that are attached to the base words to create new words. Adding derivational morphemes changes the class of the word. The derivational process of word formation in Toto is reflected by the presence of the morphemes *-paɟoa*, *-ɟoa*, *-pæva*, and *-va*.

### 3.1.2.1. Adjective to Verb

| Adjectives In English | Derivation In English | Derived Verbs in English | IPA | Roman |
|---|---|---|---|---|
| Bright | Bright-en | Brighten | /haipaɟoa/ | *Hahipajowa* |
| Short | Short-en | Shorten | /edaŋpaɟoa/ | *Edangpajowa* |
| Wide | Wide-en | Widen | /ʈaboɟoa/ | *Tabojowa* |
| White | White-en | Whiten | /haŋpapaɟoa/ | *Hangpapajowa* |
| Black | Black-en | Blacken | /daʃipaɬɑ/ | *Dashipajowa* |
| Red | Red-en | Redden | /æluipaɬɑ/ | *Alupajowa* |
| Dark | Dark-eN | Darken | /dileŋpaɬɑ/ | *Dilenpajowa* |
| Light | Light-en | Lighten | /peleŋpaɬɑ//ɦaipaɬɑ/ | *pelungpajowa/hahipajowa* |

*3.1.2.2. Verb to Noun*

| Verb in English | Derivation in English | Derived Noun in English | IPA | Roman |
|---|---|---|---|---|
| Walk | Walk-er | Walker | /təiʋɑ/ | *Taiwa* |
| Eat | Eat-er | Eater | /cɑʋɑ/ | *Chawa* |
| Play | Play-er | Player | /kəlaiʋɑ/ | *Klaina* |
| Read | Read-er | Reader | /pəraiʋɑ/ | *Paraiwa* |
| Write | Write-er | Writer | /laʋɑ/ | *Lawa* |
| Run | Run-er | Runner | /ʈʰuiʋɑ/ | *Thuiwa* |
| Talk | Talk-er | Talker | /jɔʋɑ/ | *Yhoewa* |

*3.1.2.3. Noun to Adjective*

| Noun in English | Derived Adjective in English | IPA | Roman |
|---|---|---|---|
| Attention | Attentive | /hiɲʋa koiʋɑ/ | *hingwa koiwa* |
| Anger | Angry | /ʃedaŋʋa/ | *Sedangwa* |
| Belief | Believable | /loɑ/ | *Loa* |
| Love | Lovely | /ɬeɬoŋɡʋa/ | *Jejengwa* |
| Courage | Courageous | /mucuiɲʋa/ | *Muchuingwa* |
| Danger | Dangerous | /kʰətrɑbiʋa/ | *Khatraniwa* |
| Dear | Dear | /bæro/ | *Baroh* |
| Misery | Miserable | /ləmʈuɑ/ | *Lemtua* |
| Habit | Habitual | /iuŋʋa/ | *Huingwa* |
| Insult | Insulting | /tuŋsiŋ tuicpɑʋɑ/ | *tungsing tuchpawa* |

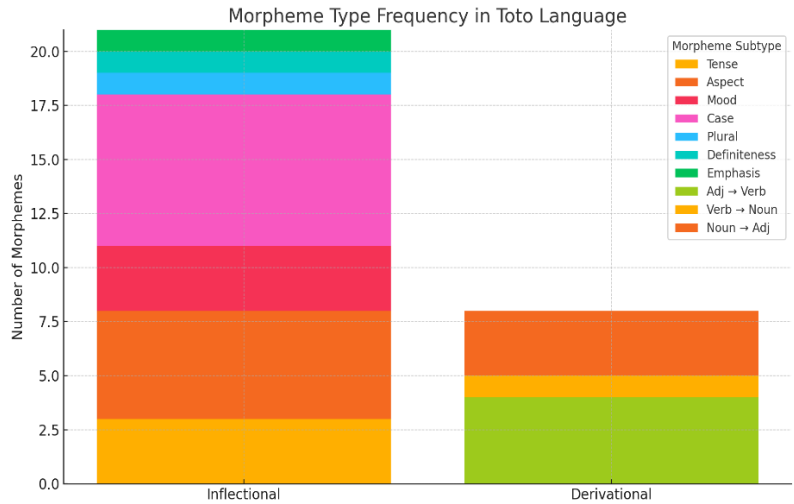

**Figure 2:** Morpheme Type Frequency in Toto

**4. AI Integration Framework: Building a Small Language Model and Trilingual Translator for Toto**
To ensure practical preservation and revitalization of the critically endangered Toto language, this project incorporates a focused Artificial Intelligence (AI) framework that emphasizes the development of a Small Language Model (SLM) alongside a Toto–Bangla–English trilingual translation engine. Unlike large-scale multilingual language models, which require extensive data and computational resources, the proposed solution is tailored to handle the limited resource availability and community-specific nature of the Toto language. The integration is structured into five key stages: corpus collection, script standardization, data processing, language model training, and application deployment.

*4.1. Corpus Collection*
A foundational step involves the collection of Toto language data through direct fieldwork in Totopara. Structured interviews, spontaneous dialogues, and cultural expressions (greetings, oral narratives, commands, and domain-specific phrases) were recorded from native speakers. Each Toto utterance was translated into Bangla and English, generating a trilingual parallel corpus. All entries were manually verified for semantic alignment by native speakers and linguists. The final corpus was stored in JSON or TSV format, maintaining the structure:

*{ "toto": "...", "bangla": "...", "english": "..." }*

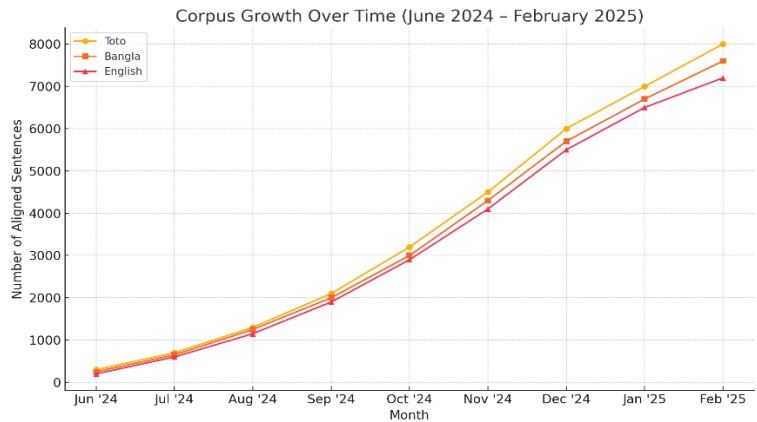

**Figure 3:** Corpus growth over time

Each entry was annotated with morpheme-level tags, part-of-speech, and syntactic boundaries for linguistic richness.

*4.2. Script Standardization*
Given the relatively recent formalization of the Toto script (2015), it was imperative to ensure its Unicode compatibility. Custom mapping tables were constructed to normalize multi-character graphemes and to support Unicode rendering. Additionally, a Romanized-to-Script transliteration tool was developed using phoneme-grapheme alignment rules. This supports both native script input and Roman script fallback for digital literacy accessibility. A lightweight keyboard interface was created using open-source tools such as Keyman or Google Input Tool APIs to facilitate script typing.

*4.3. Data Processing*
To prepare the corpus for machine learning, all textual data underwent linguistic preprocessing. This included sentence tokenization, morpheme segmentation, normalization of scripts, and validation of trilingual sentence alignment. Data augmentation techniques such as synonym substitution, conjugation expansion, and reordering were employed to artificially expand the dataset. The final corpus was divided into training, validation, and test sets, formatted as:

*Toto_sentence<TAB>Bangla_sentence<TAB>English_sentence*

The corpus also enabled the extraction of domain-specific lexicons and sentence templates to guide translation modelling.

*4.4. Small Language Model Training*
A Small Language Model (SLM) was trained specifically for the Toto language using Masked Language Modeling (MLM) objectives. A custom tokenizer was built using SentencePiece, enabling subword-level tokenization suitable for morphologically rich and low-resource languages. The architecture was a distilled Transformer model (2–4 layers, 128–256 hidden units, ~5M parameters), trained on ~20,000 Toto sentences. The SLM enables core linguistic tasks such as word prediction, phrase suggestion, and syntax-based generation.

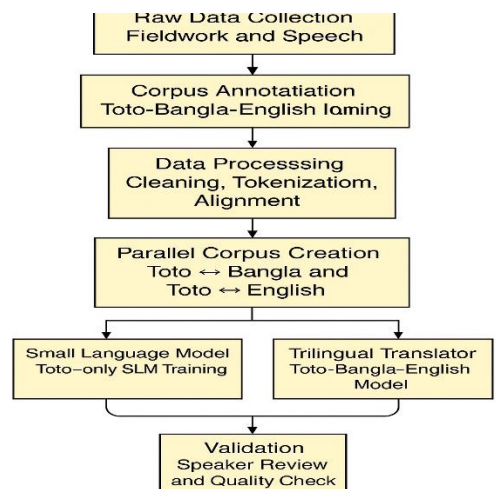

**Figure 4:** Integration of language documentation and Artificial Intelligence

*4.4.1. Toto Language Model*
A Small Language Model (SLM) was trained specifically for the Toto language using Masked Language Modeling (MLM) objectives. A custom tokenizer was built using SentencePiece, enabling subword-level tokenization suitable for morphologically rich and low-resource languages. The architecture was a distilled Transformer model (2–4 layers, 128–256 hidden units, ~5M parameters), trained on ~20,000 Toto sentences. The SLM enables core linguistic tasks such as word prediction, phrase suggestion, and syntax-based generation.

*4.4.2. Trilingual Translator (Toto–Bangla–English)*
For translation, a transformer-based encoder-decoder model was trained using the trilingual corpus. Open-source frameworks such as MarianMT, OpenNMT, or Fairseq were used with multilingual sentence-pair inputs. Language-specific control tags (e.g., <2bn>, <2en>) guided output generation. Shared subword vocabularies ensured alignment across the three languages. Fine-tuning was performed on 5,000–10,000 aligned sentences, leveraging pretrained multilingual models like mBART, T5, or distilled NLLB for initialization. Evaluation was done using BLEU, chrF, and human acceptability scores.

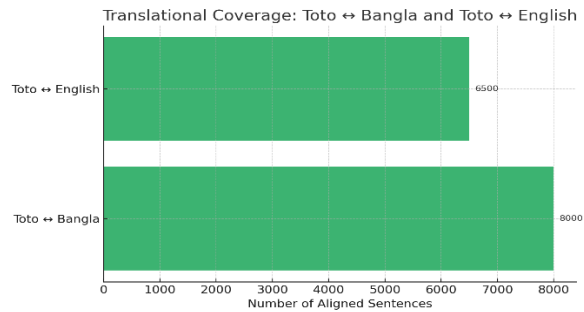

**Figure 5:** Translational coverage (Toto-Bangla & Toto-English)

*4.5. Application Deployment and Use*
The final AI models were integrated into a lightweight web and mobile application. The app supports:
Toto-Bangla-English translation
Morpheme-level explanation and glossing
Script display and transliteration
Offline inference using ONNX or TensorFlow Lite for accessibility in low-connectivity environments

**5. Challenges and Limitations**

*5.1. Linguistic Challenges*
Number of Speakers: With speaker strength being 1606, data collection and linguistic verification has been difficult, as the variables are limited in number.
Communication: Until the development of the Toto script in 2015, the language was primarily transmitted orally. This poses challenges in the written documentation of the language, which impacts the Digital Archiving process as well.
Variation: The language shows variations in pronunciation and structure across generations, complicating the standardization process required for digital documentation.

*5.2. Technological Challenges*
Data Limitations for AI Training: The limited number of orthographic materials in Toto constrains the accuracy of the model, as AI models require extensive training data.

This research aims to address these challenges and develop a sustainable, ethically sound, and technologically feasible approach to language preservation.

*5.3. Discussion and Implications*
This study demonstrates the potential of AI in language preservation, providing a structured morphological analysis of the Toto language while integrating modern technology for practical applications. The documentation of grammatical structures enhances linguistic understanding, while the AI-powered trilingual learning model ensures accessibility and usability.

*5.4. Key Contributions*
This research provides a comprehensive analysis of Toto's inflectional and derivational morphology.

*5.5. AI-Based Language Learning Model*
The project integrates AI-driven translation tools, making the language more accessible.
The Toto-Bangla-English parallel corpus facilitates language learning and digital preservation.

*5.6. Cultural and Educational Impact*
Toto- speakers can strengthen their mother tongue while simultaneously focusing on learning English, and Bengali, ensuring multilingual competence, and also Non-native learners (researchers, educators, tourists) can engage with the language, promoting cross-cultural understanding.

**6. Conclusions**
The preservation of endangered languages is a critical area of research, especially in the face of globalization and language shift. This study focuses on the Toto language, a critically endangered tongue spoken by a small community in West Bengal, and presents a linguistic analysis alongside an AI-driven solution for its preservation. By documenting inflectional and derivational morphemes, this research provides a comprehensive grammatical framework for the language, which serves as a pivotal step in its documentation and revitalization.

A major contribution of this research is the development of an AI-based trilingual learning application, which integrates Toto, Bangla, and English to facilitate multilingual education while ensuring that the Toto language remains relevant in modern contexts. By leveraging artificial intelligence and natural language processing (NLP), this application provides an interactive, scalable, and sustainable tool for both native Toto speakers and non-native learners. The parallel corpus created in this study forms the foundation of AI training, ensuring accurate translation, morphology processing, and linguistic representation of the language. Beyond technological advancements, this research also highlights the sociolinguistic factors influencing language shift, such as intergenerational language transmission, community engagement, and educational policies. The AI-based model is designed to complement oral traditions rather than replace them, ensuring that the natural linguistic identity of the Toto people is preserved while adapting to the digital age.

Furthermore, this study contributes to broader discussions on linguistic diversity, digital archiving, and indigenous language education. By integrating traditional linguistic fieldwork with computational modelling, it provides a blueprint for future endangered language preservation projects, bridging the gap between theory and practical application. The involvement of local speakers, linguists, and AI researchers ensures that this initiative is ethically sound and culturally sensitive. The findings underscore the importance of community participation, government support, and academic collaboration in sustaining indigenous languages.

**Acknowledgements**
From the bottom of our hearts, we acknowledge the support of the Indian Council of Social Science Research (ICSSR) for funding this ongoing project on the documentation and analysis of the Toto language, of which the present research forms a significant part. We extend our heartfelt thanks to the members of the Toto community, especially Dhaniram Toto, Bharat Toto, Asish Toto, Sone Toto, Bhakta Toto, and Bhabesh Toto, for their generous participation and for providing rich linguistic data crucial to our analysis. Their time, insights, and commitment to preserving their language have been invaluable. This work would not have been possible without their trust and collaboration. We are deeply appreciative of their contributions, which continue to shape the trajectory of our research. We also thank our institutional partners and colleagues for their ongoing support. We hope that this work contributes meaningfully to the preservation and understanding of the endangered Toto language.

Guha, *et.al*, 2025  *Waiguo Yuyan yu Wenhua*